\documentclass[11pt,a4paper]{article}
\usepackage[nohyperref]{acl2019}
\usepackage[hyperindex,breaklinks]{hyperref} 
\usepackage{times}
\usepackage{latexsym}
\usepackage{linguex}
\usepackage{amsmath}
\usepackage{amssymb}
\usepackage{booktabs}
\usepackage{linguex}
\usepackage{url}
\usepackage{array}
\newcolumntype{x}[1]{>{\centering\arraybackslash\hspace{0pt}}p{#1}}
\newcolumntype{C}[1]{>{\centering\let\newline\\\arraybackslash\hspace{0pt}}m{#1}}
\usepackage{tabularx}
\usepackage{graphicx}
\usepackage{dblfloatfix}
\usepackage{enumitem}
\usepackage{ textcomp }
\usepackage{todonotes}
\usepackage{subcaption}

\usepackage{url}

\aclfinalcopy 

\title{Right for the Wrong Reasons: Diagnosing Syntactic Heuristics in \\ Natural Language Inference}

\author{R. Thomas McCoy,\textsuperscript{1} Ellie Pavlick,\textsuperscript{2} \& Tal Linzen\textsuperscript{1}  \\
\textsuperscript{1}Department of Cognitive Science, Johns Hopkins University\\
\textsuperscript{2}Department of Computer Science, Brown University\\
\texttt{tom.mccoy@jhu.edu}, \texttt{ellie\_pavlick@brown.edu}, \texttt{tal.linzen@jhu.edu} 
}

\date{}

\begin{document}
\maketitle
\begin{abstract}
    A machine learning system can 
    score well on a given test set by relying on heuristics that are effective for frequent example types but break down in more challenging cases. We study this issue within natural language inference (NLI), the task of determining whether one sentence entails another. We hypothesize that statistical NLI models may adopt three fallible syntactic heuristics: the lexical overlap heuristic, the subsequence heuristic, and the constituent heuristic. To determine whether models have adopted these heuristics, we introduce a controlled evaluation set called HANS (Heuristic Analysis for NLI Systems), which contains many examples where the heuristics fail. We find that models trained on MNLI, including BERT, a state-of-the-art model, perform very poorly on HANS, suggesting that they have indeed adopted these heuristics. We conclude that there is substantial room for improvement in NLI systems, and that the HANS dataset can motivate and measure progress in this area. 
\end{abstract}

\setlength{\Exlabelwidth}{0.25em}
\setlength{\SubExleftmargin}{1.35em}

\begin{table*}
\resizebox{\textwidth}{!}{
\begin{tabular}{lp{6.5cm}p{6cm}} \toprule
    Heuristic & Definition & \multicolumn{1}{l}{Example}  \\ \midrule
    Lexical overlap & Assume that a premise entails all hypotheses constructed from words in the premise & \textbf{The doctor} was \textbf{paid} by \textbf{the actor}. \newline $\xrightarrow[\textsc{WRONG}]{}$ The doctor paid the actor. \\ 
    \midrule
    Subsequence & Assume that a premise entails all of its contiguous subsequences. & The doctor near \textbf{the actor danced}. \newline $\xrightarrow[\textsc{WRONG}]{}$ The actor danced. \\
     \midrule
    Constituent & Assume that a premise entails all complete subtrees in its parse tree. & If \textbf{the artist slept}, the actor ran. \newline  $\xrightarrow[\textsc{WRONG}]{}$ The artist slept.\\
    \bottomrule
\end{tabular}
}
\caption{The heuristics targeted by the HANS dataset, along with examples of incorrect entailment predictions that these heuristics would lead to.} \label{tab:intro_examples}
\end{table*}

\section{Introduction}

Neural networks excel at learning the statistical patterns in a training set and applying them to test cases drawn from the same distribution as the training examples. 
This strength can also be a weakness: statistical learners such as standard neural network architectures are prone to adopting shallow heuristics that succeed for the majority of training examples, instead of learning the underlying generalizations that they are intended to capture. If such heuristics often yield correct outputs, the loss function provides little incentive for the model 
to learn to generalize to more challenging cases as a human performing the task would. 

This issue has been documented across domains in artificial intelligence. In computer vision, for example, 
neural networks trained to recognize objects are misled by contextual heuristics: a network that is able to recognize monkeys in a typical context with high accuracy may nevertheless label a monkey holding a guitar as a human, since in the training set guitars tend to co-occur with humans but not monkeys 
\cite{wang2018visual}. Similar heuristics arise in visual question answering systems \cite{agarwal2016analyzing}.

The current paper addresses this issue in the domain of natural language inference (NLI), the task of determining whether a \textbf{premise} sentence entails (i.e., implies the truth of) a \textbf{hypothesis} sentence \cite{condoravdi2003entailment, dagan2005pascal,bowman2015large}. As in other domains, neural NLI models have been shown to learn shallow heuristics, in this case based on the presence of specific words \cite{naik2018stress, sanchez2018behavior}. For example, a model might assign a label of \textit{contradiction} to any input containing the word \textit{not}, since \textit{not} often appears in the examples of contradiction in standard NLI training sets.

The focus of our work is on heuristics that are based on superficial \textbf{syntactic} properties.
Consider the following sentence pair, which has the target label \textit{entailment}:

\ex. \label{ex:entcorrect}\textit{Premise:} The judge was paid by the actor.\\
\textit{Hypothesis:} The actor paid the judge.

An NLI system that labels this example correctly might do so not by reasoning about the meanings of these sentences, but rather by assuming that the premise entails any hypothesis whose words all appear in the premise \cite{dasgupta2018evaluating,naik2018stress}.
Crucially, if the model is using this heuristic, it will predict \textit{entailment} for \ref{ex:passive_incorrect} as well, even though that label is incorrect in this case:

\ex. \label{ex:passive_incorrect} \textit{Premise:} The actor was paid by the judge.\\
\textit{Hypothesis:} The actor paid the judge.

\noindent
We introduce a new evaluation set called HANS (Heuristic Analysis for NLI Systems), 
designed to diagnose the use of such fallible structural heuristics.\footnote{GitHub repository with data and code: \url{https://github.com/tommccoy1/hans}} We target three heuristics, defined in Table~\ref{tab:intro_examples}.
While these heuristics often yield correct labels, they are not valid inference strategies because they fail on many examples.
We design our dataset around such examples, so that models that employ these heuristics are guaranteed to fail on particular subsets of the dataset, rather than simply show lower overall accuracy. 

We evaluate four popular NLI models, including BERT, a state-of-the-art model \cite{devlin2019bert}, on the HANS dataset. All models performed substantially below chance on this dataset, barely exceeding 0\% accuracy in most cases. We conclude that their behavior is consistent with the hypothesis that they have adopted these heuristics.

\paragraph{Contributions:} This paper has three main contributions. First, we introduce the HANS dataset, an NLI evaluation set that tests specific hypotheses about invalid heuristics that NLI models are likely to learn. Second, we use this dataset to illuminate interpretable shortcomings in state-of-the-art models trained on MNLI \cite{williams2018multinli}; these shortcoming may arise from inappropriate model inductive biases, from insufficient signal provided by training datasets, or both. Third, we show that these shortcomings can be made less severe by augmenting a model's training set with the types of examples present in HANS. 
These results indicate that there is substantial room for improvement for current NLI models and datasets, and that HANS can serve as a tool for motivating and measuring progress in this area.

\begin{table*}
\resizebox{\textwidth}{!}{
\begin{tabular}{lllc} \toprule
  Heuristic & Premise & Hypothesis & Label \\ \midrule
  Lexical & The banker near the judge saw the actor. & The banker saw the actor. &  E  \\
  overlap & The lawyer was advised by the actor. & The actor advised the lawyer. & E  \\
  heuristic & The doctors visited the lawyer. & The lawyer visited the doctors. & N \\
  & The judge by the actor stopped the banker. & The banker stopped the actor. & N \\
  \midrule
  Subsequence  & The artist and the student called the judge. & The student called the judge. & E \\
  heuristic & Angry tourists helped the lawyer. & Tourists helped the lawyer. & E \\
  & The judges heard the actors resigned. & The judges heard the actors. & N \\
  & The senator near the lawyer danced. & The lawyer danced. & N \\
  \midrule
  Constituent & Before the actor slept, the senator ran. & The actor slept. & E  \\
  heuristic & The lawyer knew that the judges shouted. & The judges shouted. & E \\
  & If the actor slept, the judge saw the artist. & The actor slept. & N \\
  & The lawyers resigned, or the artist slept. & The artist slept. & N \\
  \bottomrule
\end{tabular}
}
\caption{Examples of sentences used to test the three heuristics. The \textit{label} column shows the correct label for the sentence pair; \textit{E} stands for \textit{entailment} and \textit{N} stands for \textit{non-entailment}. A model relying on the heuristics would label all examples as \textit{entailment} (incorrectly for those marked as N).} \label{tab:heur_examples}
\end{table*}

\section{Syntactic Heuristics}

We focus on three heuristics: the lexical overlap heuristic, the subsequence heuristic, and the constituent heuristic, all defined in Table~\ref{tab:intro_examples}. 
These heuristics form a hierarchy: the constituent heuristic is a special case of the subsequence heuristic, which in turn is a special case of the lexical overlap heuristic. Table~\ref{tab:heur_examples} in the next page gives examples where each heuristic succeeds and fails.

There are two reasons why we expect these heuristics to be adopted by a statistical learner trained on standard NLI training datasets such as SNLI \cite{bowman2015large} or MNLI \cite{williams2018multinli}. First, the MNLI training set contains far more examples that support the heuristics than examples that contradict them:\footnote{In this table, the lexical overlap counts include the subsequence counts, which include the constituent counts.}

\vspace{0.5\baselineskip}
\noindent\resizebox{\columnwidth}{!}{
\noindent\begin{tabular}{lp{2cm}p{2cm}} \toprule
    Heuristic & Supporting Cases & Contradicting Cases \\ \midrule
    Lexical overlap & 2,158 & 261 \\
    Subsequence & 1,274 & 72 \\
    Constituent & 1,004 & 58 \\ \bottomrule
\end{tabular}
}
\vspace{0.5\baselineskip}

\noindent 
Even the 261 contradicting cases in MNLI may not provide strong evidence against the heuristics.
For example, 133 of these cases contain negation in the premise but not the hypothesis, as in \ref{ex:negcontra}. Instead of using these cases to overrule the lexical overlap heuristic, a model might account for them by learning to assume that the label is \textit{contradiction} whenever there is negation in the premise but not the hypothesis \cite{mccoy2019scil}:

\ex. \label{ex:negcontra} 
\a. I \textbf{don't} care. $\nrightarrow$ I care.
\b. This is \textbf{not} a contradiction. $\nrightarrow$ This is a contradiction.

\noindent There are some examples in MNLI that contradict the heuristics in ways that are not easily explained away by other heuristics; see Appendix \ref{app:contra} for examples. However, such cases are likely too rare to discourage a model from learning these heuristics.
MNLI contains data from multiple genres, so we conjecture that the scarcity of contradicting examples is not just a property of one genre, but rather a general property of NLI data generated in the crowdsourcing approach used for MNLI. We thus hypothesize that any crowdsourced NLI dataset would make our syntactic heuristics attractive to statistical learners without strong linguistic priors. 

The second reason we might expect current NLI models to adopt these heuristics is that their input representations 
may make them susceptible to these heuristics. 
The lexical overlap heuristic 
disregards the order of the words in the sentence and considers only their identity, 
so it is likely to be adopted by bag-of-words NLI models (e.g., \citealt{parikh2016decomp}). The subsequence heuristic considers linearly adjacent chunks of words, so one might expect it to be adopted by standard RNNs, which process sentences in linear order. Finally, the constituent heuristic appeals to components of the parse tree, so one might expect to see it adopted by tree-based NLI models \cite{bowman2016spinn}.

\section{Dataset Construction}

For each heuristic, we generated five templates for examples that support the heuristic and five templates for examples that contradict it. Below is one template for the subsequence heuristic; see Appendix~\ref{appendix:templates} for a full list of templates.

\ex. The N$_1$ P the N$_2$ V. $\nrightarrow$ The N$_2$ V. \\\textit{The lawyer by the actor ran.} $\nrightarrow$ \textit{The actor ran.}

\noindent We generated 1,000 examples from each template, for a total of 10,000 examples per heuristic. Some heuristics are special cases of others, but we made sure that the examples
for one heuristic did not also fall under a more narrowly defined heuristic. That is, for lexical overlap cases, 
the hypothesis was not a subsequence or constituent of the premise; for subsequence cases, the hypothesis was 
not a constituent of the premise.

\subsection{Dataset Controls}

\paragraph{Plausibility:} One advantage of generating examples from templates---instead of, e.g., modifying naturally-occurring examples---%
is that we can ensure the 
plausibility of all generated sentences. For example, we do not generate cases such as \textit{The student read the book} $\nrightarrow$ \textit{The book read the student}, which could ostensibly be solved using a hypothesis-plausibility heuristic. To achieve this, we drew our core vocabulary from \newcite{ettinger2018assessing}, where every noun was a plausible subject of every verb or a plausible object of every transitive verb. Some templates required expanding this core vocabulary; in those cases, we manually curated the additions to ensure plausibility.

\paragraph{Selectional criteria:} Some of our example types depend on the availability of lexically-specific verb frames. For example, \ref{ex:nps} requires awareness of the fact that \textit{believed} can take a clause (\textit{the lawyer saw the officer}) as its complement:

\ex. The doctor believed the lawyer saw the officer. $\nrightarrow$ The doctor believed the lawyer. \label{ex:nps}

It is arguably unfair to expect a model to understand this example if it had only ever encountered \textit{believe} with a noun phrase object (e.g., \textit{I believed \textbf{the man}}).
To control for this issue, we only chose verbs that appeared at least 50 times in the MNLI training set in all relevant frames.

\section{Experimental Setup}

Since HANS is designed to probe for structural heuristics, we selected three models that exemplify popular strategies for representing the input sentence: DA, a bag-of-words model; ESIM, which uses a sequential structure; and SPINN, which uses a syntactic parse tree. In addition to these three models, we included BERT, a state-of-the-art model for MNLI. The following paragraphs provide more details on these models.

\paragraph{DA:} The Decomposable Attention model (DA; \citealp{parikh2016decomp}) uses a form of attention to align words in the premise and hypothesis and to make predictions based on the aggregation of this alignment. It uses no word order information and can thus be viewed as a bag-of-words model. 

\paragraph{ESIM:} The Enhanced Sequential Inference Model (ESIM; \citealp{chen2017esim}) uses a modified 
bidirectional LSTM to encode sentences. We use the variant with a sequential encoder, rather than the tree-based Hybrid Inference Model (HIM).  

\paragraph{SPINN:} The Stack-augmented Parser-Interpreter Neural Network (SPINN; \citealp{bowman2016spinn}) is tree-based: it encodes sentences by combining phrases based on a syntactic parse. We use the SPINN-PI-NT variant, which takes a parse tree as an input (rather than learning to parse). For MNLI, we used the parses provided in the MNLI release; for HANS, we used parse templates that we created 
based on parses from the Stanford PCFG Parser 3.5.2 \cite{klein2003pcfg}, the same parser used to parse MNLI. Based on manual inspection, this parser generally provided correct parses for HANS examples.

\setlength\tabcolsep{6pt}

\begin{figure*}
\begin{subfigure}{0.32\textwidth}
    \centering
    \includegraphics[height=130px]{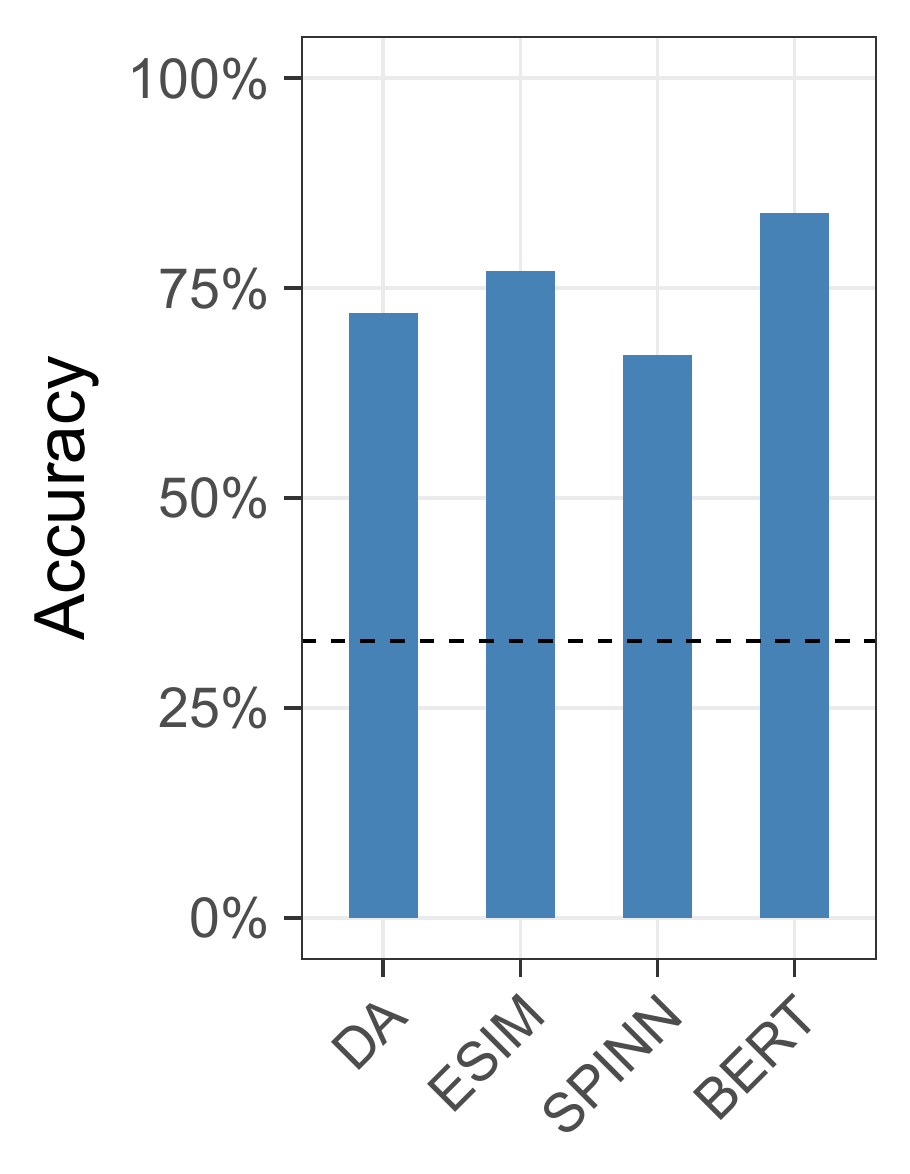}
    \caption{}
    \label{fig:mnli_accuracy}
\end{subfigure}%
\begin{subfigure}{0.68\textwidth}
    \centering
    \includegraphics[height=130px]{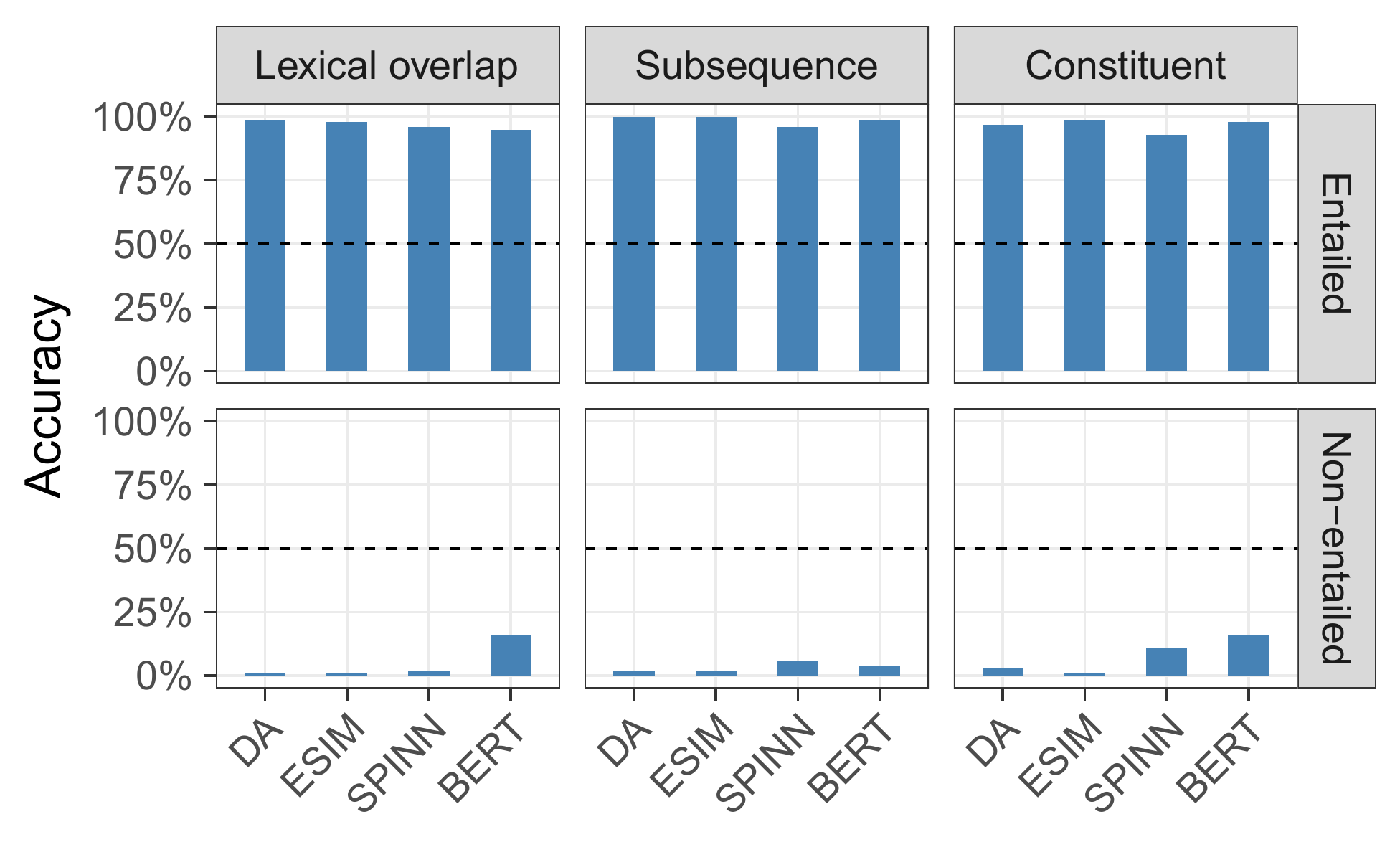}
    \caption{}
    \label{fig:hans_accuracy}
\end{subfigure}
\caption{(a) Accuracy on the MNLI test set. (b) Accuracies on the HANS evaluation set, which has six sub-components, each defined 
by its correct label
and the heuristic it addresses. Dashed lines show chance performance. All models behaved as we would expect them to if they had adopted the heuristics targeted by HANS. That is, they nearly always predicted \textit{entailment} for the examples in HANS, leading to near-perfect accuracy when the true label is \textit{entailment}, and near-zero accuracy 
when the true label is \textit{non-entailment}.
Exact results are in Appendix \ref{appendix:numerical}.
} 
\label{fig:results}
\end{figure*}

\paragraph{BERT:} The Bidirectional Encoder Representations from Transformers model (BERT; \citealp{devlin2019bert}) is a Transformer model that uses attention, rather than recurrence, to process sentences. 
We use the \texttt{bert-base-uncased} pretrained model and fine-tune it on MNLI.

\paragraph{Implementation and evaluation:} For DA and ESIM, we used the implementations from AllenNLP \citep{Gardner2017AllenNLP}. For SPINN\footnote{\url{https://github.com/stanfordnlp/spinn}; we used the NYU fork at \url{https://github.com/nyu-mll/spinn}.} and BERT,\footnote{\url{https://github.com/google-research/bert}} we used code from the GitHub repositories for the papers introducing those models.

We trained all models on MNLI. MNLI uses three labels (\textit{entailment}, \textit{contradiction}, and \textit{neutral}). We chose to annotate HANS with two labels only (\textit{entailment} and \textit{non-entailment}) because the distinction between \textit{contradiction} and \textit{neutral} was often unclear for our cases.\footnote{For example, with \textit{The actor was helped by the judge} $\nrightarrow$ \textit{The actor helped the judge}, it is possible that the actor did help the judge, pointing to a label of \textit{neutral}; yet the premise does pragmatically imply that the actor did not help the judge, meaning that this pair could also fit the non-strict definition of \textit{contradiction} used in NLI annotation.} For evaluating a model on HANS, we took the highest-scoring label  
out of \textit{entailment}, \textit{contradiction}, and \textit{neutral}; we then translated \textit{contradiction} or \textit{neutral} labels to \textit{non-entailment}. An alternate approach would have been to add the \textit{contradiction} and \textit{neutral} scores to determine a score for \textit{non-entailment}; we found little difference between these approaches, since the models almost always assigned more than 50\% of the label probability to a single label.\footnote{We also tried training the models on MNLI with \textit{neutral} and \textit{contradiction} collapsed into \textit{non-entailment}; this gave similar results as collapsing after training (Appendix \ref{appendix:merged}) .} 

\section{Results}

All models achieved high scores on the MNLI test set (Figure~\ref{fig:mnli_accuracy}), replicating the accuracies found in past work (DA: \citealt{gururangan2018annotation};  ESIM: \citealt{williams2018multinli}; SPINN: \citealt{williams2018latent}; BERT: \citealt{devlin2019bert}).
On the HANS dataset, all models almost always assigned the correct label in the cases where the label is \textit{entailment}, i.e., where the correct answer is in line with the hypothesized heuristics. However, they all performed poorly---with accuracies less than 10\% in most cases, when chance is 50\%---on the cases where the heuristics make incorrect predictions (Figure~\ref{fig:hans_accuracy}). Thus, despite their high scores on the MNLI test set, all four models behaved in a way consistent with the use of the heuristics targeted in HANS, and not with the correct rules of inference. 

\paragraph{Comparison of models:} Both DA and ESIM had near-zero performance across all three heuristics. These models might therefore make no distinction between the three heuristics, but instead treat them all as the same phenomenon, i.e. lexical overlap. Indeed, for DA, this must be the case, as this model does not have access to word order; ESIM does in theory have access to word order information but does not appear to use it here.

SPINN had the best performance on the subsequence cases. This might be due to the tree-based nature of its input: since the subsequences targeted in these cases were explicitly chosen not to be constituents, they do not form cohesive units in SPINN's input in the way they do for sequential models. SPINN also outperformed DA and ESIM on the constituent cases, suggesting that SPINN's tree-based representations moderately helped it learn 
how specific constituents contribute to the overall sentence. Finally, SPINN did worse than the other models on constituent cases where the correct answer is \textit{entailment}. This moderately greater balance between accuracy on entailment and non-entailment cases further indicates that SPINN is less likely than the other models to assume that constituents of the premise are entailed; this harms its performance in cases where that assumption happens to lead to the correct answer.

BERT did slightly worse than SPINN on the subsequence cases, but performed noticeably less poorly than all other models at both the constituent and lexical overlap cases (though it was still far below chance). Its performance particularly stood out for the lexical overlap cases, suggesting that some of BERT's success at MNLI may be due to a greater tendency to incorporate word order information compared to other models.

\paragraph{Analysis of particular example types:} In the cases where a model's performance on a heuristic was perceptibly above zero, accuracy was not evenly spread across subcases (for case-by-case results, see Appendix \ref{appendix:fine}). For example, within the lexical overlap cases, BERT achieved 39\% accuracy on conjunction (e.g., \textit{The actor and the doctor saw the artist} $\nrightarrow$ \textit{The actor saw the doctor}) but 0\% accuracy on subject/object swap (\textit{The judge called the lawyer} $\nrightarrow$ \textit{The lawyer called the judge}). Within the constituent heuristic cases, BERT achieved 49\% accuracy at determining that a clause embedded under \textit{if} and other conditional words is not entailed (\textit{If the doctor resigned, the lawyer danced} $\nrightarrow$ \textit{The doctor resigned}), but 0\% accuracy at identifying that the clause outside of the conditional clause is also not entailed (\textit{If the doctor resigned, the lawyer danced} $\nrightarrow$ \textit{The lawyer danced}).

\section{Discussion}\label{sec:discussion}

\paragraph{Independence of heuristics:}
Though each heuristic is most closely related to one class of model (e.g., the constituent heuristic is related to tree-based models), all models failed on cases illustrating all three heuristics. This finding is unsurprising since these heuristics are closely related to each other, meaning that an NLI model may adopt all of them, even the ones not specifically 
targeting that class of model. For example, 
the subsequence and constituent heuristics are special cases of the lexical overlap heuristic, so all models can fail on cases illustrating all heuristics, because all models have access to individual words. 

Though the heuristics form a hierarchy---the constituent heuristic is a subcase of the subsequence heuristic, which is a subcase of the lexical overlap heuristic---this hierarchy does not necessarily predict the performance of our models.
For example, BERT performed worse on the subsequence heuristic than on the constituent heuristic, even though the constituent heuristic is a special case of the subsequence heuristic. Such behavior has two possible causes. First, it could be due to the specific cases we chose for each heuristic: the cases chosen for the subsequence heuristic may be inherently more challenging than the cases chosen for the constituent heuristic, even though the constituent heuristic as a whole is a subset of the subsequence one. Alternately,
it is possible for a model to adopt a more general heuristic (e.g., the subsequence heuristic) but to make an exception for some special cases (e.g., the cases to which the constituent heuristic could apply).

\paragraph{Do the heuristics arise from the architecture or the training set?} The behavior of a trained model depends on both the training set and the model's architecture. The models' poor results on HANS could therefore arise from architectural limitations, from insufficient signal in the MNLI training set, or from both.

The fact that SPINN did markedly better at the constituent and subsequence cases than ESIM and DA, even though the three models were trained on the same dataset, suggests that MNLI does contain some signal that can counteract the appeal of the syntactic heuristics tested by HANS. SPINN's structural inductive biases allow it to leverage this signal, but the other models' biases do not.

Other sources of evidence suggest that the models' failure is due in large part to insufficient signal from the MNLI training set, rather than the models' representational capacities alone. The BERT model we used (\texttt{bert-base-uncased}) was found by \newcite{goldberg2019assessing} to achieve strong results in syntactic tasks such as subject-verb agreement prediction, a task that minimally requires a distinction between the subject and direct object of a sentence \cite{linzen2016assessing,gulordava2018colorless,marvin2018targeted}. Despite this evidence that BERT has access to relevant syntactic information, its accuracy was 0\% on the subject-object swap cases (e.g., \textit{The doctor saw the lawyer} $\nrightarrow$ \textit{The lawyer saw the doctor}). We believe it is unlikely that our fine-tuning step on MNLI, a much smaller corpus than the corpus BERT was trained on, substantially changed the model's representational capabilities. Even though the model most likely had access to information about subjects and objects, then, MNLI did not make it clear how that information applies to inference. Supporting this conclusion, \newcite{mccoy2018rnns} found little evidence of compositional structure in the InferSent model, which was trained on SNLI, even though the same model type (an RNN) did learn clear compositional structure when trained on tasks that underscored the need for such structure. These results further suggest that the models' poor compositional behavior arises more because of the training set than because of model architecture.

Finally, our BERT-based model differed from the other models in that it was pretrained on a massive amount of data on a masking task and a next-sentence classification task, followed by fine-tuning on MNLI, while the other models were only trained on MNLI; we therefore cannot rule out the possibility that BERT's comparative success at HANS was due to the greater amount of data it has encountered rather than any architectural features. 

\paragraph{Is the dataset too difficult?} 
To assess the difficulty of our dataset, we obtained human judgments on a subset of HANS from 95 participants on Amazon Mechanical Turk as well as 3 expert annotators (linguists who were unfamiliar with HANS: 2 graduate students and 1 postdoctoral researcher). 
The average accuracy was 76\% for Mechanical Turk participants and 97\% for expert annotators; further details are in Appendix~\ref{appendix:crowd}.

Our Mechanical Turk results contrast with those of \newcite{nangiaconservative}, who report an accuracy of 92\% in the same population on examples from MNLI; this indicates that HANS is indeed more challenging for humans than MNLI is.
The difficulty of some of our examples is in line with past psycholinguistic work in which humans have been shown to incorrectly answer comprehension questions for some of our subsequence subcases. For example, in an experiment in which participants read the sentence \textit{As Jerry played the violin gathered dust in the attic}, some participants answered \textit{yes} to the question \textit{Did Jerry play the violin?} \cite{christianson2001thematic}. 

Crucially, although Mechanical Turk annotators found HANS to be harder overall than MNLI, their accuracy was similar whether the correct answer was \textit{entailment} (75\% accuracy) or \textit{non-entailment} (77\% accuracy). The contrast between the balance in the human errors across labels and the stark imbalance in the models' errors (Figure~\ref{fig:hans_accuracy}) indicates that human errors are unlikely to be driven by the heuristics targeted in the current work.

\section{Augmenting the training data with HANS-like examples}
\label{sec:augmentation}

The failure of the models we tested raises the question of what it would take to do well on HANS. 
One possibility is that a different type of model would perform better. For example, 
a model based on hand-coded rules might handle HANS well.
However, since most models we tested are in theory capable of handling HANS's examples but failed to do so when trained on MNLI, it is likely that performance could also be  improved by training the same architectures on a dataset in which these heuristics are less successful.

\begin{figure}
\centering
\includegraphics[width=\columnwidth]{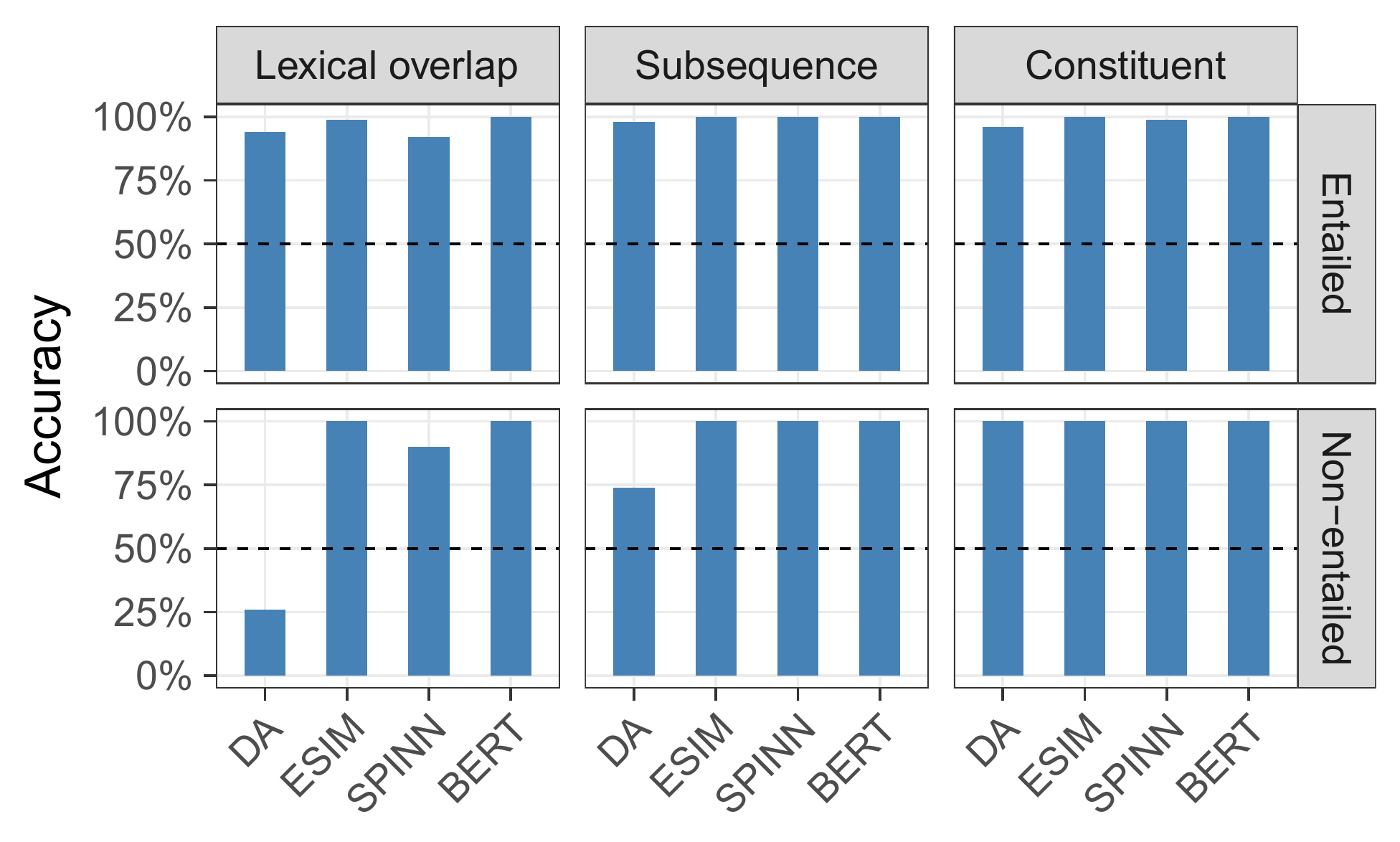}
\caption{HANS accuracies for models trained on MNLI plus examples of all 30 categories in HANS. } \label{fig:aug_results}
\end{figure}

To test that hypothesis, we retrained each model on the MNLI training set augmented with a dataset structured exactly like HANS (i.e. using the same thirty subcases) but containing no specific examples that 
appeared in HANS. Our additions comprised 30,000 examples, roughly 8\% of the size of the original MNLI training set (392,702 examples). 
In general, the models trained on the augmented MNLI 
performed very well on HANS (Figure~\ref{fig:aug_results});
the one exception 
was that the DA model performed poorly on subcases for which a bag-of-words representation was inadequate.\footnote{The effect on MNLI test set performance was less clear; the augmentation with HANS-like examples improved MNLI test set performance for BERT (84.4\% vs. 84.1\%) and ESIM (77.6\% vs 77.3\%) but hurt performance for DA (66.0\% vs. 72.4\%) and SPINN (63.9\% vs. 67.0\%).} 
This experiment is only an initial exploration and leaves open many questions about the conditions under which a model will successfully avoid a heuristic; for example, how many contradicting examples are required? At the same time, these results do suggest that, to prevent a model from learning a heuristic, one viable approach is to use a training set that does not support this heuristic.

\setlength\tabcolsep{3pt}
\setlength\tabcolsep{6pt}

\paragraph{Transfer across HANS subcases:}
The positive results of the HANS-like augmentation experiment are compatible with the possibility that the models simply memorized the templates that made up HANS's thirty subcases. To address this, we retrained our models on MNLI augmented with \textit{subsets} of the HANS cases (withholding some cases; see Appendix~\ref{appendix:withheld} for details), then tested the models on the withheld cases. 

The results of one of the transfer experiments, using BERT, are shown in Table~\ref{tab:withheld1}. 
There were some successful cases of transfer; e.g., BERT performed well on the withheld categories with sentence-initial adverbs, 
regardless of whether the correct label was \textit{non-entailment} 
or \textit{entailment}.
Such successes suggest that BERT is able to learn from some specific subcases that it should rule out the broader heuristics; in this case, the non-withheld cases plausibly informed BERT not to indiscriminately follow the constituent heuristic, encouraging it to instead base its judgments on the specific adverbs in question (e.g., \textit{certainly} vs. \textit{probably}). 
However, the models did not always transfer successfully; e.g., BERT had 0\% accuracy on entailed passive examples when such examples were withheld, likely because the training set still included many non-entailed passive examples, meaning that BERT may have learned to assume that all sentences with passive premises are cases of non-entailment.
Thus, though the models do seem to be able to rule out the broadest versions of the heuristics and transfer that knowledge to some new cases, they may still fall back to
the heuristics for other cases. 
For further results involving withheld categories, see Appendix~\ref{appendix:withheld}.

\begin{table}[t]
    \centering
    \resizebox{\columnwidth}{!}{
    \begin{tabular}{p{7cm}l} \toprule
        Withheld category &  Results \\ \midrule
        Lexical overlap: Conjunctions ($\nrightarrow$) \newline \hspace{5mm}\hangindent=5mm\textit{The doctor saw the author and the tourist.} \newline $\nrightarrow$ \textit{The author saw the tourist.} \vspace{0.15cm} & \raisebox{-0.7\totalheight}{\includegraphics[height=60px]{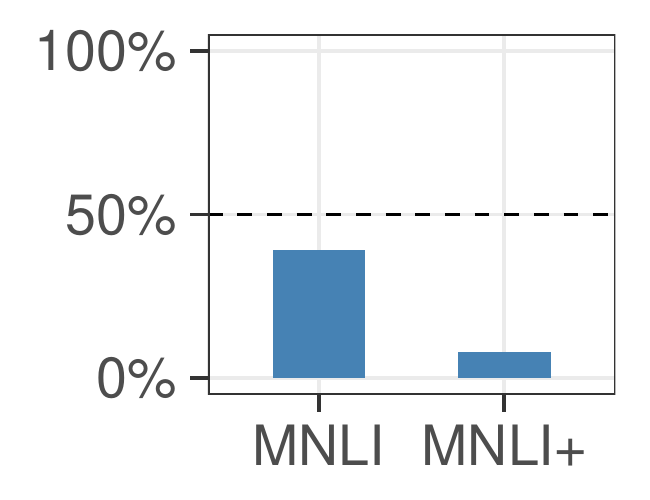}} \\

        Lexical overlap: Passives ($\rightarrow$) \newline \hspace{5mm}\hangindent=5mm\textit{The authors were helped by the actor.} \newline $\rightarrow$ \textit{The actor helped the authors.}& \raisebox{-0.7\totalheight}{\includegraphics[height=60px]{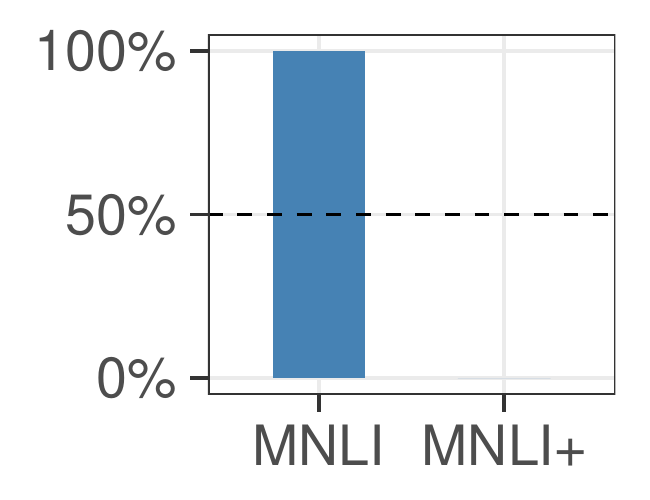}}  \\

        Subsequence: NP/Z ($\nrightarrow$) \newline \hangindent=5mm\textit{Before the actor moved the doctor arrived.} \newline $\nrightarrow$ \textit{The actor moved the doctor.}& \raisebox{-0.7\totalheight}{\includegraphics[height=60px]{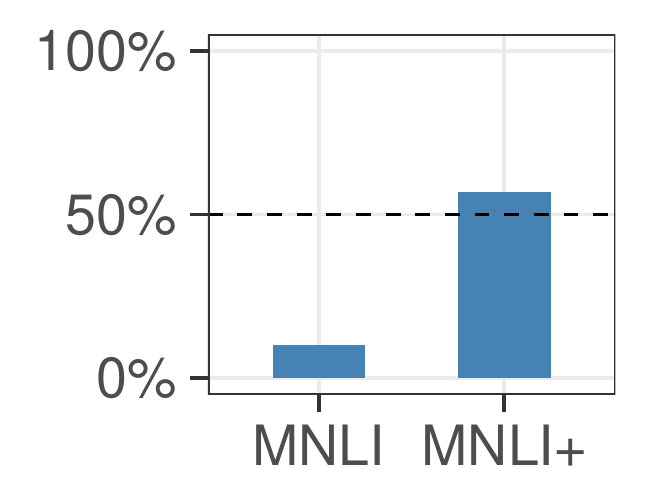}}  \\

        Subsequence: PP on object ($\rightarrow$) \newline \hangindent=5mm\textit{The authors saw the judges by the doctor.} \newline $\rightarrow$ \textit{The authors saw the judges.} & \raisebox{-0.7\totalheight}{\includegraphics[height=60px]{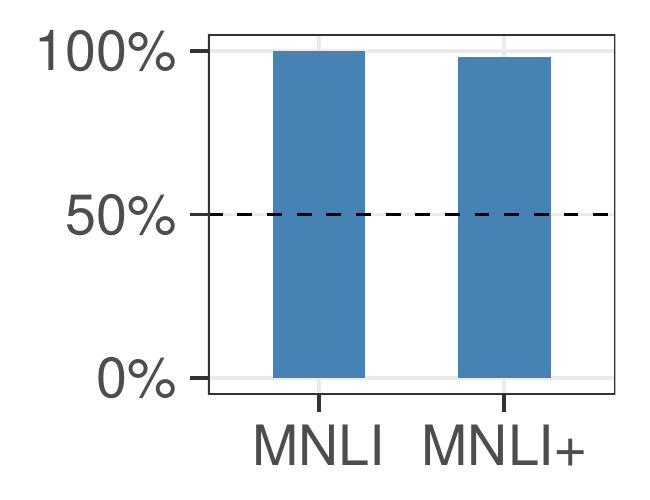}}  \\

        Constituent: Adverbs ($\nrightarrow$) \newline \hangindent=5mm\textit{Probably the artists helped the authors.} \newline $\nrightarrow$ \textit{The artists helped the authors.} & \raisebox{-0.7\totalheight}{\includegraphics[height=60px]{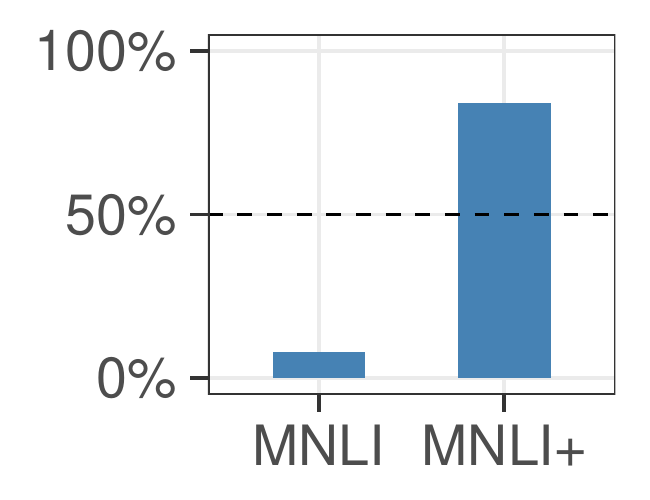}}  \\

        Constituent: Adverbs ($\rightarrow$) \newline \hangindent=5mm\textit{Certainly the lawyers shouted.} \newline $\rightarrow$ \textit{The lawyers shouted.} & \raisebox{-0.7\totalheight}{\includegraphics[height=60px]{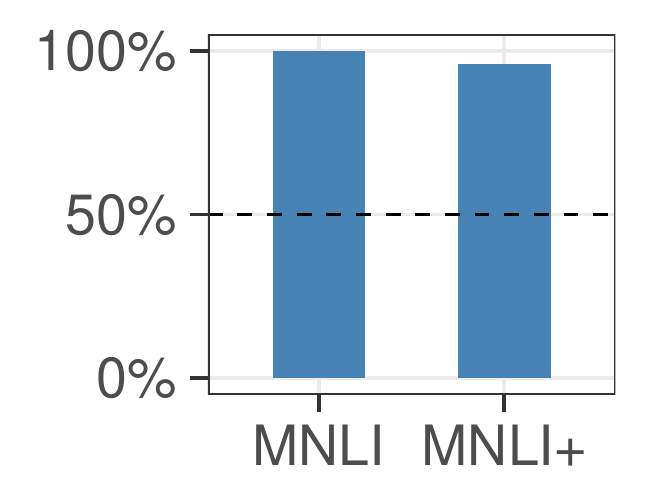}}  \\
        
        \bottomrule
    \end{tabular}
    }
    \caption{Accuracies for BERT fine-tuned on basic MNLI and on MNLI+, which is MNLI augmented with most HANS categories except withholding the categories in this table. The two lexical overlap cases shown here are adversarial in that MNLI+ contains cases superficially similar to them but with opposite labels (namely, the \textit{Conjunctions ($\rightarrow$)} and \textit{Passives ($\nrightarrow$)} cases from Table \ref{tab:lex_templates} in the Appendix). The remaining cases in this table are not adversarial in this way. 
    }
    \label{tab:withheld1}
\end{table}

\paragraph{Transfer to an external dataset:}
Finally, we tested models on the \texttt{comp\_same\_short} and \texttt{comp\_same\_long} datasets from \newcite{dasgupta2018evaluating}, which consist of lexical overlap cases:

\ex. the famous and arrogant cat is not more nasty than the dog with glasses in a white dress. $\nrightarrow$ the dog with glasses in a white dress is not more nasty than the famous and arrogant cat. \label{ex:dasguptab}

This dataset differs from HANS in at least three important ways: it is based on a phenomenon not present in HANS (namely, comparatives); it uses a different vocabulary from HANS; and many of its sentences are semantically implausible. 

We used this dataset to test both BERT fine-tuned on MNLI, and BERT fine-tuned on MNLI augmented with HANS-like examples. The augmentation improved performance modestly for the long examples and dramatically for the short examples, suggesting that training with HANS-like examples has benefits that extend beyond HANS.\footnote{We hypothesize that HANS helps more with short examples because most HANS sentences are short.}

\begin{figure}
    \centering
    \includegraphics[width=0.7\columnwidth]{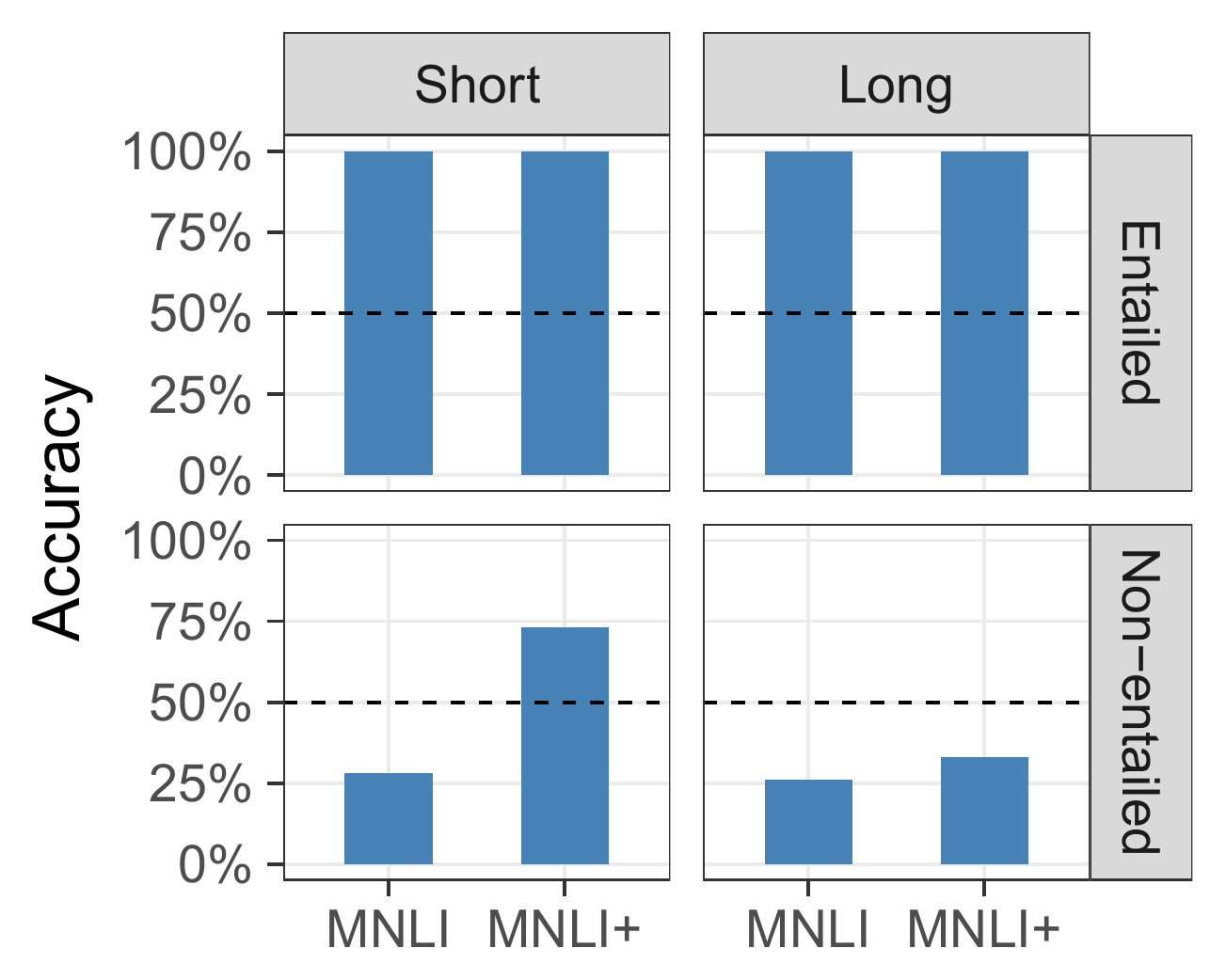}
    \caption{Results on the lexical overlap cases from \newcite{dasgupta2018evaluating} for BERT fine-tuned on MNLI or on MNLI augmented with HANS-like examples.}
    \label{fig:dasgupta}
\end{figure}

\section{Related Work}

\subsection{Analyzing trained models}

This project relates to an extensive body of research on exposing and understanding weaknesses in models' learned behavior and representations.
In the NLI literature, \newcite{poliak2018hypothesis} and \newcite{gururangan2018annotation} show that, due to biases  in NLI datasets, it is possible to achieve far better than chance accuracy on those datasets by only looking at the hypothesis.
Other recent works address possible ways in which NLI models might use fallible heuristics, focusing on semantic phenomena, such as lexical inferences \cite{glockner2018} or quantifiers \cite{geiger2018stress}, or biases based on specific words \cite{sanchez2018behavior}. Our work focuses instead on \textit{structural} phenomena, following the proof-of-concept work done by \newcite{dasgupta2018evaluating}. Our focus on using NLI to address how models capture structure follows some older work about using NLI for the evaluation of parsers \cite{rimell2010pete,mehdad2010syntactic}.

NLI has been used to investigate many other types of linguistic information besides syntactic structure \cite{poliak2018dnc, white2017everything}. Outside NLI, multiple projects have used classification tasks to understand what linguistic and/or structural information is present in vector encodings of sentences \cite[e.g.,][]{adi2017fine,ettinger2018assessing,conneau2018cram}. We instead choose the behavioral approach of using task performance on critical cases. Unlike the classification approach, this approach is agnostic to model structure;
our dataset could be used to evaluate a symbolic NLI system just as easily as a neural one, whereas typical classification approaches only work for models with vector representations.

\subsection{Structural heuristics}

Similar to our lexical overlap heuristic,  \newcite{dasgupta2018evaluating}, \newcite{nie2018analyzing}, and \newcite{kim2018teaching} also tested NLI models on specific phenomena where word order matters; we use a larger set of phenomena to study a more general notion of lexical overlap that is less dependent on the properties of a single phenomenon, such as passives.
\newcite{naik2018stress} also find evidence that NLI models use a lexical overlap heuristic, but our approach is substantially different from theirs.\footnote{\newcite{naik2018stress} diagnose the lexical overlap heuristic by appending \textit{and true is true} to existing MNLI hypotheses, which decreases lexical overlap but does not change the sentence pair's label. We instead generate new sentence pairs for which the words in the hypothesis all appear in the premise.}

This work builds on our pilot study in \newcite{mccoy2019scil}, which studied one of the  subcases of the subsequence heuristic. 
Several of our subsequence subcases are inspired by psycholinguistics research  \cite{bever1970cognitive, frazier1982making,tabor2004effects}; these works have aims similar to ours but are concerned with the representations used by 
humans rather than neural networks.

Finally, all of our constituent heuristic subcases depend on the implicational behavior of specific words. Several past works \cite{pavlick2016tense,rudinger2018factuality,white2018lexicosyntactic,white2018role} have studied such behavior for verbs (e.g., \textit{He knows it is raining} entails \textit{It is raining}, while \textit{He believes it is raining} does not). We extend that approach by including  other types of words with specific implicational behavior, namely conjunctions (\textit{and, or}), prepositions that take clausal arguments (\textit{if, because}), and adverbs (\textit{definitely, supposedly}). \newcite{maccartney2009natural} also discuss the implicational behavior of these various types of words within NLI.

\subsection{Generalization}

Our work suggests that test sets drawn from the same distribution as the training set may be inadequate for assessing whether a model has learned to perform the intended task. Instead, it is also necessary to evaluate on a generalization set that departs from the training distribution. 
\newcite{mccoy2018revisiting} found a similar result for the task of question formation; different architectures that all succeeded on the test set failed on the generalization set in different ways, showing that the test set alone was not sufficient to determine what the models had learned. This effect can arise not just from different architectures but also from different initializations of the same architecture \cite{weber2018fine}.

\section{Conclusions}

Statistical learners such as neural networks closely track the statistical regularities in their training sets. This process makes them vulnerable to adopting heuristics that are valid for frequent cases but fail on less frequent ones. We have investigated three such heuristics that we hypothesize NLI models are likely to learn. To evaluate whether NLI models do behave consistently with these heuristics, we have introduced the HANS dataset, on which models using these heuristics are guaranteed to fail. We find that four existing NLI models perform very poorly on HANS, suggesting that their high accuracies on NLI test sets may be due to the exploitation of invalid heuristics rather than deeper understanding of language. However, these models performed significantly better on both HANS and on a separate structure-dependent dataset when their training data was augmented with HANS-like examples. Overall, our results indicate that, despite the impressive accuracies of state-of-the-art models on standard evaluations, there is still much progress to be made and that targeted, challenging datasets, such as HANS, are important for determining whether models are learning what they are intended to learn.

\section*{Acknowledgments}

We are grateful to Adam Poliak, Benjamin Van Durme, Samuel Bowman, the members of the JSALT General-Purpose Sentence Representation Learning team, and the members of the Johns Hopkins Computation and Psycholinguistics Lab for helpful comments, and to Brian Leonard for assistance with the Mechanical Turk experiment. Any errors remain our own.

This material is based upon work supported by the National Science Foundation Graduate Research Fellowship Program under Grant No. 1746891 and the 2018 Jelinek Summer Workshop on Speech and Language Technology (JSALT). Any opinions, findings, and conclusions or recommendations expressed in this material are those of the authors and do not necessarily reflect the views of the National Science Foundation or the JSALT workshop.

\bibliographystyle{acl_natbib}
\bibliography{hans}

\appendix

\section{MNLI examples that contradict the HANS heuristics} \label{app:contra}

The sentences in \ref{ex:goodmnli} show examples from the MNLI training set that contradict the lexical overlap, subsequence, and constituent heuristics. 
The full set of all 261 contradicting examples in the MNLI training set may be viewed at \url{https://github.com/tommccoy1/hans/blob/master/mnli_contradicting_examples}.

\ex. \label{ex:goodmnli} \a. A subcategory of accuracy is consistency. $\nrightarrow$ Accuracy is a subcategory of consistency.
\b. At the same time, top Enron executives were free to exercise their stock options, and some did. $\nrightarrow$ Top Enron executives were free to exercise.
\c. She was chagrined at The Nation's recent publication of a column by conservative education activist Ron Unz arguing that liberal education reform has been an unmitigated failure. $\nrightarrow$ Liberal education reform has been an unmitigated failure.

\section{Templates}\label{appendix:templates}

Tables \ref{tab:lex_templates}, \ref{tab:subseq_templates}, and \ref{tab:const_templates} contain the templates for the lexical overlap heuristic, the subsequence heuristic, and the constituent heuristic, respectively.

In some cases, a given template has multiple versions, such as one version where a noun phrase modifier attaches to the subject and another where the modifier attaches to the object. For clarity, we have only listed one version of each template here. The full list of templates can be viewed in the code on GitHub.\footnote{\url{https://github.com/tommccoy1/hans}}

{
\renewcommand{\arraystretch}{1.5}
\begin{table*}
\resizebox{\textwidth}{!}{
\begin{tabular}{p{3.5cm}p{5.5cm}p{6cm}} \toprule
    Subcase & Template & Example \\ \midrule
    Entailment: \newline \raggedright Untangling relative clauses & The N$_1$ who the N$_2$ V$_1$ V$_2$ the N$_3$ \newline $\rightarrow$ The N$_2$ V$_1$ the N$_1$. & The athlete who the judges admired called the manager. \newline $\rightarrow$ The judges admired the athlete.\\
    Entailment: \newline Sentences with PPs & The N$_1$ P the N$_2$ V the N$_3$ \newline $\rightarrow$ The N$_1$ V the N$_3$ & The tourists by the actor recommended the authors. \newline $\rightarrow$ The tourists recommended the authors. \\
    Entailment:\newline \raggedright Sentences with relative clauses & The N$_1$ that V$_2$  V$_1$ the N$_2$  \newline $\rightarrow$ The N$_1$ V$_1$ the N$_2$ & The actors that danced saw the author.  \newline $\rightarrow$ The actors saw the author. \\
    Entailment:\newline Conjunctions & The N$_1$ V the N$_2$ and the N$_3$ \newline $\rightarrow$ The N$_1$ V the N$_3$ & The secretaries encouraged the scientists and the actors. \newline $\rightarrow$ The secretaries encouraged the actors.\\
    Entailment:\newline Passives & The N$_1$ were V by the N$_2$ \newline $\rightarrow$ The N$_1$ V the N$_2$ & The authors were supported by the tourists. \newline $\rightarrow$ The tourists supported the authors. \\ \midrule
    Non-entailment:\newline Subject-object swap &  The N$_1$ V the N$_2$. \newline $\nrightarrow$ The N$_2$ V the N$_1$. & The senators mentioned the artist. \newline $\nrightarrow$ The artist mentioned the senators. \\
    Non-entailment:\newline Sentences with PPs & The N$_1$ P the N$_2$ V the N$_3$ \newline $\nrightarrow$ The N$_3$ V the N$_2$ & The judge behind the manager saw the doctors. \newline $\nrightarrow$ The doctors saw the manager. \\
    Non-entailment:\newline \raggedright Sentences with relative clauses & The N$_1$ V$_1$ the N$_2$ who the N$_3$ V$_2$ \newline $\nrightarrow$ The N$_2$ V$_1$ the N$_3$ & The actors advised the manager who the tourists saw. \newline $\nrightarrow$ The manager advised the tourists. \\
    Non-entailment:\newline Conjunctions & The N$_1$ V the N$_2$ and the N$_3$ \newline $\nrightarrow$ The N$_2$ V the N$_3$ & The doctors advised the presidents and the tourists. \newline $\nrightarrow$ The presidents advised the tourists. \\
    Non-entailment:\newline Passives & The N$_1$ were V by the N$_2$ \newline $\nrightarrow$ The N$_1$ V the N$_2$ & The senators were recommended by the managers. \newline $\nrightarrow$ The senators recommended the managers. \\ \bottomrule
\end{tabular}
}
\caption{Templates for the lexical overlap heuristic}  \label{tab:lex_templates}
\end{table*}
}

{
\renewcommand{\arraystretch}{1.5}
\begin{table*}
\resizebox{\textwidth}{!}{
\begin{tabular}{p{4cm}p{5cm}p{6cm}} \toprule
    Subcase & Template & Example \\ \midrule
    Entailment:\newline Conjunctions & The N$_1$ and the N$_2$ V the N$_3$ \newline $\rightarrow$ The N$_2$ V the N$_3$ & The actor and the professor mentioned the lawyer. \newline $\rightarrow$ The professor mentioned the lawyer. \\
    Entailment:\newline Adjectives & Adj N$_1$ V the N$_2$ \newline $\rightarrow$ N$_1$ V the N$_2$ & Happy professors mentioned the lawyer. \newline $\rightarrow$ Professors mentioned the lawyer. \\
    Entailment:\newline Understood argument & The N$_1$ V the N$_2$ \newline $\rightarrow$ The N$_1$ V & The author read the book. \newline $\rightarrow$ The author read. \\
    Entailment:\newline Relative clause on object & The N$_1$ V$_1$ the N$_2$ that V$_2$ the N$_3$ \newline $\rightarrow$ The N$_1$ V$_1$ the N$_2$ & The artists avoided the senators that thanked the tourists. \newline $\rightarrow$ The artists avoided the senators. \\
    Entailment: \newline PP on object & The N$_1$ V the N$_2$ P the N$_3$ \newline $\rightarrow$ The N$_1$ V the N$_2$ & The authors supported the judges in front of the doctor. \newline $\rightarrow$ The authors supported the judges. \\ \midrule
    Non-entailment:\newline NP/S & The N$_1$ V$_1$ the N$_2$ V$_2$ the N$_3$ \newline $\nrightarrow$ The N$_1$ V$_1$ the N$_2$ & The managers heard the secretary encouraged the author. \newline $\nrightarrow$ The managers heard the secretary. \\
    Non-entailment:\newline PP on subject & The N$_1$ P the N$_2$ V \newline $\nrightarrow$ The N$_2$ V &  The managers near the scientist resigned. \newline $\nrightarrow$ The scientist resigned. \\
    Non-entailment:\newline Relative clause on subject & The N$_1$ that V$_1$ the N$_2$ V$_2$ the N$_3$ \newline $\nrightarrow$ The N$_2$ V$_2$ the N$_3$ & The secretary that admired the senator saw the actor. \newline $\nrightarrow$ The senator saw the actor. \\
    Non-entailment:\newline MV/RR & The N$_1$ V$_1$ P the N$_2$ V$_2$   \newline $\nrightarrow$ The N$_1$ V$_1$ P the N$_2$ & The senators paid in the office danced. \newline $\nrightarrow$ The senators paid in the office. \\
    Non-entailment:\newline NP/Z & P the N$_1$ V$_1$ the N$_2$ V$_2$ the N$_3$ \newline $\nrightarrow$ The N$_1$ V$_1$ the N$_2$ & Before the actors presented the professors advised the manager. \newline $\nrightarrow$ The actors presented the professors. \\ \bottomrule
\end{tabular}
}
\caption{Templates for the subsequence heuristic} \label{tab:subseq_templates}
\end{table*}
}

{
\renewcommand{\arraystretch}{1.5}
\begin{table*}
\resizebox{\textwidth}{!}{
\begin{tabular}{p{4cm}p{5cm}p{6cm}} \toprule
    Subcase & Template & Example \\ \midrule
    Entailment:\newline Embedded under preposition & P the N$_1$ V$_1$, the N$_2$ V$_2$ the N$_3$ \newline $\rightarrow$ The N$_1$ V$_1$ & Because the banker ran, the doctors saw the professors. \newline $\rightarrow$ The banker ran.\\
    Entailment:\newline Outside embedded clause & P the N$_1$ V$_1$ the N$_2$, the N$_3$ V$_2$ the N$_4$ \newline $\rightarrow$ The N$_3$ V$_2$ the N$_4$ & Although the secretaries recommended the managers, the judges supported the scientist. \newline $\rightarrow$ The judges supported the scientist.\\
    Entailment:\newline Embedded under  verb & The N$_1$ V$_1$ that the N$_2$ V$_2$ \newline $\rightarrow$ The N$_2$ V$_2$ & The president remembered that the actors performed. \newline $\rightarrow$ The actors performed. \\
    Entailment:\newline Conjunction & The N$_1$ V$_1$, and the N$_2$ V$_2$ the N$_3$. \newline $\rightarrow$ The N$_2$ V$_2$ the N$_3$ & The lawyer danced, and the judge supported the doctors. \newline $\rightarrow$ The judge supported the doctors. \\
    Entailment:\newline Adverbs & Adv the N V \newline $\rightarrow$ The N V & Certainly the lawyers resigned. \newline $\rightarrow$ The lawyers resigned. \\ \midrule
    Non-entailment:\newline Embedded under  preposition & P the N$_1$ V$_1$, the N$_2$ V$_2$ the N$_2$ \newline $\nrightarrow$ The N$_1$ V$_1$ & Unless the senators ran, the professors recommended the doctor. \newline $\nrightarrow$ The senators ran. \\
    Non-entailment:\newline Outside embedded clause & P the N$_1$ V$_1$ the N$_2$, the N$_3$ V$_2$ the N$_4$ \newline $\nrightarrow$ The N$_3$ V$_2$ the N$_4$ & Unless the authors saw the students, the doctors helped the bankers. \newline $\nrightarrow$ The doctors helped the bankers. \\
    Non-entailment:\newline Embedded under verb & The N$_1$ V$_1$ that the N$_2$ V$_2$ the N$_3$ \newline $\nrightarrow$ The N$_2$ V$_2$ the N$_3$ & The tourists said that the lawyer saw the banker. \newline $\nrightarrow$ The lawyer saw the banker. \\
    Non-entailment:\newline Disjunction & The N$_1$ V$_1$, or the N$_2$ V$_2$ the N$_3$ \newline $\nrightarrow$ The N$_2$ V$_2$ the N$_3$ & The judges resigned, or the athletes mentioned the author. \newline $\nrightarrow$ The athletes mentioned the author.\\
    Non-entailment:\newline Adverbs & Adv the N$_1$ V the N$_2$ \newline $\nrightarrow$ The N$_1$ V the N$_2$ & Probably the artists saw the authors. \newline $\nrightarrow$ The artists saw the authors.\\ \bottomrule
\end{tabular}
}
\caption{Templates for the constituent heuristic} \label{tab:const_templates}
\end{table*}
}

\section{Fine-grained results} \label{appendix:fine}

Table \ref{tab:fine_grained_entailed} shows the results by subcase for models trained on MNLI for the subcases where the correct answer is \textit{entailment}. Table \ref{tab:fine_grained_nonentailed} shows the results by subcase for these models for the subcases where the correct answer is \textit{non-entailment}.

\begin{table*}
\centering
\resizebox{\textwidth}{!}{
\begin{tabular}{p{2cm}p{5cm}ccccc} \toprule
     Heuristic & Subcase & DA & ESIM & SPINN & BERT \\ \midrule
     Lexical & Untangling relative clauses & 0.97 & 0.95 & 0.88 & 0.98 \\
     overlap & \multicolumn{6}{l}{\textit{The athlete who the judges saw called the manager.} $\rightarrow$ \textit{The judges saw the athlete.}} \\
     \\
     & Sentences with PPs & 1.00 & 1.00 & 1.00 & 1.00 \\
     & \multicolumn{6}{l}{\textit{The tourists by the actor called the authors.}  $\rightarrow$ \textit{The tourists called the authors.}}\\ \\
     & Sentences with relative clauses & 0.98 & 0.97 & 0.97 & 0.99 \\
     & \multicolumn{6}{l}{\textit{The actors that danced encouraged the author.}  $\rightarrow$ \textit{The actors encouraged the author.}} \\ \\
     & Conjunctions & 1.00 & 1.00 & 1.00 & 0.77 \\
     & \multicolumn{6}{l}{\textit{The secretaries saw the scientists and the actors.} $\rightarrow$ \textit{The secretaries saw the actors.}}\\ \\
     & Passives & 1.00 & 1.00 & 0.95 & 1.00  \\ 
     & \multicolumn{6}{l}{\textit{The authors were supported by the tourists.} $\rightarrow$ \textit{The tourists supported the authors.}}\\ \\
     \midrule
     Subsequence & Conjunctions & 1.00 & 1.00 & 1.00 & 0.98 \\
     & \multicolumn{6}{l}{\textit{The actor and the professor shouted.} $\rightarrow$ \textit{The professor shouted.}}\\ \\
     & Adjectives & 1.00 & 1.00 & 1.00 & 1.00 \\
     & \multicolumn{6}{l}{\textit{Happy professors mentioned the lawyer.} $\rightarrow$ \textit{Professors mentioned the lawyer.}}\\ \\
     & Understood argument & 1.00 & 1.00 & 0.84 & 1.00 \\ 
     & \multicolumn{6}{l}{\textit{The author read the book.} $\rightarrow$ \textit{The author read.}} \\ \\
     & Relative clause on object & 0.98 & 0.99 & 0.95 & 0.99 \\ 
     & \multicolumn{6}{l}{\textit{The artists avoided the actors that performed.} $\rightarrow$ \textit{The artists avoided the actors.}}\\ \\
     & PP on object & 1.00 & 1.00 & 1.00 & 1.00 \\ 
     & \multicolumn{6}{l}{\textit{The authors called the judges near the doctor.} $\rightarrow$ \textit{The authors called the judges.}}\\ \\
     \midrule
     Constituent & Embedded under preposition & 0.99 & 0.99 & 0.85 & 1.00 \\ 
     & \multicolumn{6}{l}{\textit{Because the banker ran, the doctors saw the professors.} $\rightarrow$ \textit{The banker ran.}}\\ \\
     & Outside embedded clause & 0.94 & 1.00 & 0.95 & 1.00 \\
     & \multicolumn{6}{l}{\textit{Although the secretaries slept, the judges danced.} $\rightarrow$ \textit{The judges danced.}}\\ \\
     & Embedded under verb & 0.92 & 0.94 & 0.99 & 0.99 \\
     & \multicolumn{6}{l}{\textit{The president remembered that the actors performed.} $\rightarrow$ \textit{The actors performed.}}\\ \\
     & Conjunction & 0.99 & 1.00 & 0.89 & 1.00 \\ 
     & \multicolumn{6}{l}{\textit{The lawyer danced, and the judge supported the doctors.} $\rightarrow$ \textit{The lawyer danced.}}\\ \\
     & Adverbs & 1.00 & 1.00 & 0.98 & 1.00 \\
     & \multicolumn{6}{l}{\textit{Certainly the lawyers advised the manager.}  $\rightarrow$ \textit{The lawyers advised the manager.}}\\ \\
     \bottomrule
\end{tabular}
}
\caption{Results for the subcases where the correct label is \textit{entailment}.} \label{tab:fine_grained_entailed}
\end{table*}

\begin{table*}
\centering
\resizebox{\textwidth}{!}{
\begin{tabular}{p{2cm}p{5cm}cccccc} \toprule
     Heuristic & Subcase & DA & ESIM & SPINN & BERT \\ \midrule
     Lexical & Subject-object swap & 0.00 & 0.00 & 0.03 & 0.00 \\
     overlap & \multicolumn{6}{l}{\textit{The senators mentioned the artist.}  $\nrightarrow$ \textit{The artist mentioned the senators.}}\\ \\
      & Sentences with PPs & 0.00 & 0.00 & 0.01 & 0.25 \\
     & \multicolumn{6}{l}{\textit{The judge behind the manager saw the doctors.}  $\nrightarrow$ \textit{The doctors saw the manager.}}\\ \\
     & Sentences with relative clauses & 0.04 & 0.04 & 0.06 & 0.18 \\
     & \multicolumn{6}{l}{\textit{The actors called the banker who the tourists saw.} $\nrightarrow$ \textit{The banker called the tourists.}}\\ \\
     & Conjunctions & 0.00 & 0.00 & 0.01 & 0.39 \\
     & \multicolumn{6}{l}{\textit{The doctors saw the presidents and the tourists.} $\nrightarrow$ \textit{The presidents saw the tourists.}}\\ \\
     & Passives & 0.00 & 0.00 & 0.00 & 0.00 \\ 
     & \multicolumn{6}{l}{\textit{The senators were helped by the managers.} $\nrightarrow$ \textit{The senators helped the managers.}}\\ \\
     \midrule
     Subsequence & NP/S & 0.04 & 0.02 & 0.09 & 0.02 \\
     & \multicolumn{6}{l}{\textit{The managers heard the secretary resigned.} $\nrightarrow$ \textit{The managers heard the secretary.}}\\ \\ 
     & PP on subject & 0.00 & 0.00 & 0.00 & 0.06 \\
     & \multicolumn{6}{l}{\textit{The managers near the scientist shouted.} $\nrightarrow$ \textit{The scientist shouted.}}\\ \\
     & Relative clause on subject & 0.03 & 0.04 & 0.05 & 0.01 \\
     & \multicolumn{6}{l}{\textit{The secretary that admired the senator saw the actor.} $\nrightarrow$ \textit{The senator saw the actor.}}\\ \\
     & MV/RR & 0.04 & 0.03 & 0.03 & 0.00 & \\
     & \multicolumn{6}{l}{\textit{The senators paid in the office danced.} $\nrightarrow$ \textit{The senators paid in the office.}}\\ \\
     & NP/Z & 0.02 & 0.01 & 0.11 & 0.10 \\ 
     & \multicolumn{6}{l}{\textit{Before the actors presented the doctors arrived.} $\nrightarrow$ \textit{The actors presented the doctors.}}\\ \\
     \midrule
     Constituent & Embedded under preposition & 0.14 & 0.02 & 0.29 & 0.50 \\
     & \multicolumn{6}{l}{\textit{Unless the senators ran, the professors recommended the doctor.}  $\nrightarrow$ \textit{The senators ran.}}\\ \\
     & Outside embedded clause & 0.01 & 0.00 & 0.02 & 0.00 \\
     & \multicolumn{6}{l}{\textit{Unless the authors saw the students, the doctors resigned.} $\nrightarrow$ \textit{The doctors resigned.}}\\ \\
     & Embedded under verb & 0.00 & 0.00 & 0.01 & 0.22 \\
     & \multicolumn{6}{l}{\textit{The tourists said that the lawyer saw the banker.} $\nrightarrow$ \textit{The lawyer saw the banker.}}\\ \\
     & Disjunction & 0.01 & 0.03 & 0.20 & 0.01 \\
     & \multicolumn{6}{l}{\textit{The judges resigned, or the athletes saw the author.} $\nrightarrow$ \textit{The athletes saw the author.}}\\ \\
     & Adverbs & 0.00 & 0.00 & 0.00 & 0.08 \\ 
     & \multicolumn{6}{l}{\textit{Probably the artists saw the authors.}  $\nrightarrow$ \textit{The artists saw the authors.}}\\ \\
     \bottomrule
\end{tabular}
}
\caption{Results for the subcases where the correct label is \textit{non-entailment}.} \label{tab:fine_grained_nonentailed}
\end{table*}

\section{Results for models trained on MNLI with \textit{neutral} and \textit{contradiction} merged} \label{appendix:merged}

Table \ref{tab:results_merged} shows the results on HANS for models trained on MNLI with the labels \textit{neutral} and \textit{contradiction} merged in the training set into the single label \textit{non-entailment}. The results are similar to the results obtained by merging the labels after training, with the models generally outputting \textit{entailment} for all HANS examples, whether that was the correct answer or not.

\setlength\tabcolsep{5pt}
\begin{table*}[h]
\centering
\begin{tabular}{llcccccc}\toprule
&  & \multicolumn{3}{c}{Correct: \textit{Entailment}} & \multicolumn{3}{c}{Correct: \textit{Non-entailment}}\\
\cmidrule(lr){3-5} \cmidrule(lr){6-8}

Model & Model class & Lexical & Subseq. & Const. & Lexical & Subseq. & Const.   \\ \midrule
DA & Bag-of-words & 1.00 & 1.00 & 0.98 & 0.00 & 0.00 & 0.03 \\
ESIM & RNN & 0.99 & 1.00 & 1.00 & 0.00 & 0.01 & 0.00 \\
SPINN & TreeRNN & 0.94 & 0.96 & 0.93 & 0.06 & 0.14 & 0.11 \\
BERT & Transformer & 0.98 & 1.00 & 0.99 & 0.04 & 0.02 & 0.20\\ 
\bottomrule
\end{tabular}
\caption{Results for models trained on MNLI with \textit{neutral} and \textit{contradiction} merged into a single label, \textit{non-entailment}. } \label{tab:results_merged}
\end{table*}
\setlength\tabcolsep{6pt}

\section{Results with augmented training with some subcases withheld} \label{appendix:withheld}

For each model, we ran five experiments, each one having 6 of the 30 subcases withheld. Each trained model was then evaluated on the categories that had been withheld from it. The results of these experiments are in Tables \ref{tab:fine_grained_entailed_withheld1}, \ref{tab:fine_grained_entailed_withheld2}, \ref{tab:fine_grained_entailed_withheld3}, \ref{tab:fine_grained_entailed_withheld4} and \ref{tab:fine_grained_entailed_withheld5}.

\begin{table*}
\centering
\resizebox{\textwidth}{!}{
\begin{tabular}{p{2cm}p{5cm}cccccc} \toprule
     Heuristic & Subcase & DA & ESIM & SPINN & BERT \\ \midrule
     Lexical & Subject-object swap & 0.01 & 1.00 & 1.00 & 1.00 \\
     overlap & \multicolumn{7}{l}{\textit{The senators mentioned the artist.}  $\nrightarrow$ \textit{The artist mentioned the senators.}}\\ \\
     Lexical & Untangling relative clauses & 0.34 & 0.23 & 0.23 & 0.20 \\
     overlap & \multicolumn{7}{l}{\textit{The athlete who the judges saw called the manager.} $\rightarrow$ \textit{The judges saw the athlete.}} \\
     \\
     Subsequence & NP/S & 0.27 & 0.00 & 0.00 & 0.10 \\
     & \multicolumn{7}{l}{\textit{The managers heard the secretary resigned.} $\nrightarrow$ \textit{The managers heard the secretary.}}\\ \\ 
     Subsequence & Conjunctions & 0.49 & 0.38 & 0.38 & 0.38 \\
     & \multicolumn{7}{l}{\textit{The actor and the professor shouted.} $\rightarrow$ \textit{The professor shouted.}}\\ \\
     Constituent & Embedded under preposition & 0.51 & 0.51 & 0.51 & 1.00 \\
     & \multicolumn{7}{l}{\textit{Unless the senators ran, the professors recommended the doctor.}  $\nrightarrow$ \textit{The senators ran.}}\\ \\
     Constituent &  Embedded under preposition & 1.00 & 0.06 & 1.00 & 0.03 \\ 
     & \multicolumn{7}{l}{\textit{Because the banker ran, the doctors saw the professors.} $\rightarrow$ \textit{The banker ran.}}\\ \\
     \bottomrule
\end{tabular}
}
\caption{Accuracies for models trained on MNLI augmented with most HANS example categories except withholding the categories in this table (experiment 1/5 for the withheld category investigation).} \label{tab:fine_grained_entailed_withheld1}
\end{table*}

\begin{table*}
\centering
\resizebox{\textwidth}{!}{
\begin{tabular}{p{2cm}p{5cm}cccccc} \toprule
     Heuristic & Subcase & DA & ESIM & SPINN & BERT \\ \midrule
     Lexical & Sentences with PPs & 0.00 & 0.96 & 0.71 & 0.97 \\
     overlap & \multicolumn{7}{l}{\textit{The judge behind the manager saw the doctors.}  $\nrightarrow$ \textit{The doctors saw the manager.}}\\ \\
     Lexical & Sentences with PPs & 1.00 & 1.00 & 0.94 & 1.00 \\
     overlap & \multicolumn{7}{l}{\textit{The tourists by the actor called the authors.}  $\rightarrow$ \textit{The tourists called the authors.}}\\ \\
     Subsequence & PP on subject & 0.00 & 0.07 & 0.57 & 0.39 \\
     & \multicolumn{7}{l}{\textit{The managers near the scientist shouted.} $\nrightarrow$ \textit{The scientist shouted.}}\\ \\
     Subsequence & Adjectives & 0.71 & 0.99 & 0.64 & 1.00 \\
     & \multicolumn{7}{l}{\textit{Happy professors mentioned the lawyer.} $\rightarrow$ \textit{Professors mentioned the lawyer.}}\\ \\
     Constituent & Outside embedded clause & 0.78 & 1.00 & 1.00 & 0.17 \\
     & \multicolumn{7}{l}{\textit{Unless the authors saw the students, the doctors resigned.} $\nrightarrow$ \textit{The doctors resigned.}}\\ \\
     Constituent & Outside embedded clause & 0.78 & 0.78 & 0.78 & 0.97 \\
     & \multicolumn{7}{l}{\textit{Although the secretaries slept, the judges danced.} $\rightarrow$ \textit{The judges danced.}}\\ \\
     \bottomrule
\end{tabular}
}
\caption{Accuracies for models trained on MNLI augmented with most HANS example categories except withholding the categories in this table (experiment 2/5 for the withheld category investigation).} \label{tab:fine_grained_entailed_withheld2}
\end{table*}

\begin{table*}
\centering
\resizebox{\textwidth}{!}{
\begin{tabular}{p{2cm}p{5cm}cccccc} \toprule
     Heuristic & Subcase & DA & ESIM & SPINN & BERT \\ \midrule
     Lexical & Sentences with relative clauses & 0.00 & 0.04 & 0.02 & 0.84 \\
     overlap & \multicolumn{7}{l}{\textit{The actors called the banker who the tourists saw.} $\nrightarrow$ \textit{The banker called the tourists.}}\\ \\
     Lexical & Sentences with relative clauses & 1.00 & 0.97 & 1.00 & 1.00 \\
     overlap & \multicolumn{7}{l}{\textit{The actors that danced encouraged the author.}  $\rightarrow$ \textit{The actors encouraged the author.}} \\ \\
     Subsequence & Relative clause on subject & 0.00 & 0.04 & 0.00 & 0.93 \\
     & \multicolumn{7}{l}{\textit{The secretary that admired the senator saw the actor.} $\nrightarrow$ \textit{The senator saw the actor.}}\\ \\
     Subsequence & Understood argument & 0.28 & 1.00 & 0.81 & 0.94 \\ 
     & \multicolumn{7}{l}{\textit{The author read the book.} $\rightarrow$ \textit{The author read.}} \\ \\
     Constituent & Embedded under verb & 0.00 & 0.00 & 0.05 & 0.98 \\
     & \multicolumn{7}{l}{\textit{The tourists said that the lawyer saw the banker.} $\nrightarrow$ \textit{The lawyer saw the banker.}}\\ \\
     Constituent & Embedded under verb & 1.00 & 0.94 & 0.98 & 0.43 \\
     & \multicolumn{7}{l}{\textit{The president remembered that the actors performed.} $\rightarrow$ \textit{The actors performed.}}\\ \\
     \bottomrule
\end{tabular}
}
\caption{Accuracies for models trained on MNLI augmented with most HANS example categories except withholding the categories in this table (experiment 3/5 for the withheld category investigation).} \label{tab:fine_grained_entailed_withheld3}
\end{table*}

\begin{table*}
\centering
\resizebox{\textwidth}{!}{
\begin{tabular}{p{2cm}p{5cm}cccccc} \toprule
     Heuristic & Subcase & DA & ESIM & SPINN & BERT \\ \midrule
     Lexical & Passives & 0.00 & 0.00 & 0.00 & 0.00 \\ 
     overlap & \multicolumn{7}{l}{\textit{The senators were helped by the managers.} $\nrightarrow$ \textit{The senators helped the managers.}}\\ \\
     Lexical & Conjunctions & 0.05 & 0.51 & 0.52 & 1.00 \\
     overlap & \multicolumn{7}{l}{\textit{The secretaries saw the scientists and the actors.} $\rightarrow$ \textit{The secretaries saw the actors.}}\\ \\
     Subsequence & MV/RR & 0.76 & 0.44 & 0.32 & 0.07 & \\
     & \multicolumn{7}{l}{\textit{The senators paid in the office danced.} $\nrightarrow$ \textit{The senators paid in the office.}}\\ \\
     Subsequence & Relative clause on object & 0.72 & 1.00 & 0.99 & 0.99 \\ 
     & \multicolumn{7}{l}{\textit{The artists avoided the actors that performed.} $\rightarrow$ \textit{The artists avoided the actors.}}\\ \\
     Constituent & Disjunction & 0.11 & 0.29 & 0.51 & 0.44 \\
     & \multicolumn{7}{l}{\textit{The judges resigned, or the athletes saw the author.} $\nrightarrow$ \textit{The athletes saw the author.}}\\ \\
     Constituent & Conjunction & 0.99 & 1.00 & 0.74 & 1.00 \\ 
     & \multicolumn{7}{l}{\textit{The lawyer danced, and the judge supported the doctors.} $\rightarrow$ \textit{The lawyer danced.}}\\ \\
     \bottomrule
\end{tabular}
}
\caption{Accuracies for models trained on MNLI augmented with most HANS example categories except withholding the categories in this table (experiment 4/5 for the withheld category investigation).} \label{tab:fine_grained_entailed_withheld4}
\end{table*}

\begin{table*}
\centering
\resizebox{\textwidth}{!}{
\begin{tabular}{p{2cm}p{5cm}cccccc} \toprule
     Heuristic & Subcase & DA & ESIM & SPINN & BERT \\ \midrule
     Lexical & Conjunctions & 0.00 & 0.44 & 0.00 & 0.08 \\
     overlap & \multicolumn{7}{l}{\textit{The doctors saw the presidents and the tourists.} $\nrightarrow$ \textit{The presidents saw the tourists.}}\\ \\
     Lexical & Passives & 0.00 & 0.00 & 0.00 & 0.00  \\ 
     overlap & \multicolumn{7}{l}{\textit{The authors were supported by the tourists.} $\rightarrow$ \textit{The tourists supported the authors.}}\\ \\
     Subsequence & NP/Z & 0.00 & 0.10 & 0.18 & 0.57 \\ 
     & \multicolumn{7}{l}{\textit{Before the actors presented the doctors arrived.} $\nrightarrow$ \textit{The actors presented the doctors.}}\\ \\
     Subsequence & PP on object & 0.04 & 0.76 & 0.04 & 0.98 \\ 
     & \multicolumn{7}{l}{\textit{The authors called the judges near the doctor.} $\rightarrow$ \textit{The authors called the judges.}}\\ \\
     Constituent & Adverbs & 0.76 & 0.33 & 0.20 & 0.84 \\ 
     & \multicolumn{7}{l}{\textit{Probably the artists saw the authors.}  $\nrightarrow$ \textit{The artists saw the authors.}}\\ \\
     Constituent & Adverbs & 0.66 & 1.00 & 0.59 & 0.96 \\
     & \multicolumn{7}{l}{\textit{Certainly the lawyers advised the manager.}  $\rightarrow$ \textit{The lawyers advised the manager.}}\\ \\
     \bottomrule
\end{tabular}
}
\caption{Accuracies for models trained on MNLI augmented with most HANS example categories except withholding the categories in this table (experiment 5/5 for the withheld category investigation).} \label{tab:fine_grained_entailed_withheld5}
\end{table*}

\section{Human experiments} \label{appendix:crowd}

To obtain human results, we used Amazon Mechanical Turk. We subdivided HANS into 114 different categories of examples, covering all possible variations of the template used to generate the example and the specific word around which the template was built. For example, for the constituent heuristic subcase of clauses embedded under verbs (e.g. \textit{The doctor believed the lawyer danced} $\nrightarrow$ \textit{The lawyer danced}), each possible verb under which the clause could be embedded (e.g. \textit{believed}, \textit{thought}, or \textit{assumed}) counted as a different category.

For each of these 114 categories, we chose 20 examples from HANS and obtained judgments from 5 human participants for each of those 20 examples. Each participant provided judgments for 57 examples plus 10 controls (67 stimuli total) and was paid \$2.00. 
The controls consisted of 5 examples where the premise and hypothesis were the same (e.g. \textit{The doctor saw the lawyer} $\rightarrow$ \textit{The doctor saw the lawyer}) and 5 examples of simple negation (e.g. \textit{The doctor saw the lawyer} $\nrightarrow$ \textit{The doctor did not see the lawyer}). 
For analyzing the data, we discarded any participants who answered any of these controls incorrectly; this led to 95 participants being retained and 105 being rejected (participants were still paid regardless of whether they were retained or filtered out). 
On average, each participant spent 6.5 seconds per example; the participants we retained spent 8.9 seconds per example, while the participants we discarded spent 4.2 seconds per example. 
The total amount of time from a participant accepting the experiment to completing the experiment averaged 17.6 minutes. This included 9.1 minutes answering the prompts (6.4 minutes for discarded participants and 12.1 minutes for retained participants) and roughly one minute spent between prompts (1 second after each prompt). The remaining time was spent reading the consent form, reading the instructions, or waiting to start (Mechanical Turk participants often wait several minutes between accepting an experiment and beginning the experiment).

The expert annotators were three native English speakers who had a background in linguistics but who had not heard about this project before providing judgments. Two of them were graduate students and one was a postdoctoral researcher. Each expert annotator labeled 124 examples (one example from each of the 114 categories, plus 10 controls).

\section{Numerical results} \label{appendix:numerical}

To facilitate future comparisons to our results, here we provide the numerical results underlying the bar plots in the main body of the paper. Table \ref{tab:resultsnum} corresponds to Figure \ref{fig:results}; the MNLI column in Table \ref{tab:resultsnum} corresponds to Figure \ref{fig:mnli_accuracy}, and the remaining columns correspond to Figure \ref{fig:hans_accuracy}. Table \ref{tab:aug_resultsnum} corresponds to Figure \ref{fig:aug_results}. The plots in Table \ref{tab:withheld1} use the numbers from the BERT columns in Tables \ref{tab:fine_grained_entailed}, \ref{tab:fine_grained_nonentailed}, and \ref{tab:fine_grained_entailed_withheld5}. Finally, the bar plots in Figure \ref{fig:dasgupta} correspond to the numerical results in Table \ref{tab:dasguptanum}.

\setlength\tabcolsep{5pt}
\begin{table*}[h]
\centering
\begin{tabular}{llp{1.5cm}cccccc}\toprule
& & & \multicolumn{3}{c}{Correct: \textit{Entailment}} & \multicolumn{3}{c}{Correct: \textit{Non-entailment}}\\
\cmidrule(lr){4-6} \cmidrule(lr){7-9}

Model & Model class & \multicolumn{1}{c}{MNLI} & Lexical & Subseq. & Const. & Lexical & Subseq. & Const.   \\ \midrule
DA & Bag-of-words & \multicolumn{1}{c}{0.72} & 0.99 & 1.00 & 0.97 & 0.01 & 0.02 & 0.03 \\
ESIM & RNN & \multicolumn{1}{c}{0.77} & 0.98 & 1.00 & 0.99 & 0.01 & 0.02 & 0.01 \\
SPINN & TreeRNN & \multicolumn{1}{c}{0.67} & 0.96 & 0.96 & 0.93 & 0.02 & 0.06 & 0.11 \\
BERT & Transformer & \multicolumn{1}{c}{0.84} & 0.95 & 0.99 & 0.98 & 0.16 & 0.04 & 0.16\\ \bottomrule
\end{tabular}
\caption{Numerical results. The MNLI column reports accuracy on the MNLI test set. The remaining columns report accuracies on 6 sub-components of the HANS evaluation set; each sub-component is defined by its correct label (either \textit{entailment} or \textit{non-entailment}) and the heuristic it addresses. } \label{tab:resultsnum}
\end{table*}
\setlength\tabcolsep{6pt}

\setlength\tabcolsep{3pt}
\begin{table}
\centering
\resizebox{\columnwidth}{!}{
\begin{tabular}{lcccccc}\toprule
& \multicolumn{3}{c}{Correct: $\rightarrow$} & \multicolumn{3}{c}{Correct: $\nrightarrow$}\\
\cmidrule(lr){2-4} \cmidrule(lr){5-7}

Model & Lex. & Subseq. & Const. & Lex. & Subseq. & Const.   \\ \midrule
DA & 0.94 & 0.98 & 0.96 & 0.26 & 0.74 & 1.00 \\
ESIM & 0.99 & 1.00 & 1.00 & 1.00 & 1.00 & 1.00 \\
SPINN & 0.92 & 1.00 & 0.99 & 0.90 & 1.00 & 1.00 \\
BERT & 1.00 & 1.00 & 1.00 & 1.00 & 1.00 & 1.00\\ 
\bottomrule
\end{tabular}
}
\caption{HANS accuracies for models trained on MNLI plus examples of all 30 categories in HANS. } \label{tab:aug_resultsnum}
\end{table}
\setlength\tabcolsep{6pt}

\begin{table}[]
    \centering
    \resizebox{\columnwidth}{!}{
    \begin{tabular}{ccccc} \toprule
            & \multicolumn{2}{c}{Correct: $\rightarrow$} & \multicolumn{2}{c}{Correct: $\nrightarrow$} \\
            
            \cmidrule(lr){2-3} \cmidrule(lr){4-5}
            
      Model & Short & Long & Short & Long  \\ \midrule
       BERT (MNLI)  & 1.00 & 1.00 & 0.28 & 0.26  \\
       BERT (MNLI+) & 1.00 & 1.00 & 0.73 & 0.33 \\ \bottomrule
    \end{tabular}
    }
    \caption{Results on the lexical overlap cases from \newcite{dasgupta2018evaluating} for BERT fine-tuned on MNLI or on MNLI augmented with HANS-like examples.}
    \label{tab:dasguptanum}
\end{table}

\end{document}